\documentclass{article}
\usepackage{spconf,amsmath,graphicx}

\usepackage{times}
\usepackage{latexsym}
\usepackage{CJK}
\usepackage{url}
\usepackage{graphicx}
\usepackage{indentfirst}
\usepackage{multirow}
\usepackage{comment}
\usepackage{amsmath}
\usepackage{xcolor}
\usepackage{multirow}
\usepackage{amssymb}
\usepackage{booktabs}
\usepackage{bm}
\usepackage{floatrow}

\newfloatcommand{capbtabbox}{table}[][\FBwidth]

\title{Integrating Subgraph-aware Relation and Direction Reasoning \\ for Question Answering}
\name{ Xu Wang\textsuperscript{\rm 1}\thanks{Contact emails: 
\{wxx,zhaoshuaiby\}@bupt.edu.cn}, Shuai Zhao\textsuperscript{\rm 1}, Bo Cheng\textsuperscript{\rm 1}, Jiale Han\textsuperscript{\rm 1}, Li Yingting\textsuperscript{\rm 1}, Hao Yang\textsuperscript{\rm 2}, Ivan Sekulic\textsuperscript{\rm 3} and Guoshun Nan\textsuperscript{\rm 4}
}
\address{\textsuperscript{\rm 1}State Key Laboratory of networking and switching technology, \\Beijing University of Posts and Telecommunications, Beijing, China\\ 
 \textsuperscript{\rm 2} 2012 Labs, Huawei Technologies CO., LTD, Beijing, China \\
 \textsuperscript{\rm 3}University of Lugano \textsuperscript{\rm 4}
 StatNLP Research Group, Singapore University of Technology and Design}

\begin{document}
%
\maketitle

\begin{abstract}
	
	Question Answering (QA) models over Knowledge
	Bases (KBs) are capable of providing more precise answers
	by utilizing relation information among entities.
	Although
	effective, most of these models solely rely on fixed
	relation representations to obtain answers for
	different question-related KB subgraphs. Hence,   
	the rich structured information of these subgraphs may be overlooked by the relation representation vectors.
	Meanwhile, the direction information of reasoning, which has been proven effective for the answer prediction on graphs,  has not been fully explored in existing work.
	To address these challenges, we propose a novel neural model, Relation-updated Direction-guided Answer Selector (RDAS), which converts relations in each subgraph to additional nodes to learn structure information. Additionally, we utilize direction information to enhance the reasoning ability. Experimental results show that our model yields substantial improvements on two widely used datasets.

\end{abstract}

\section{Introduction}

Knowledge bases have become the critical resources in a variety of natural language processing applications. A KB such as Freebase \cite{DBLP:conf/sigmod/BollackerEPST08}, always contains millions of facts which are composed of subject-predicate-object triples, also referred to as a relation between two entities. Such rich structured information has proven effective in KB-based Question Answering (KBQA) tasks which aim to find the single answer (or multiple answers) to a factoid question using facts in the targeting KB \cite{survey}.

Early studies on KBQA leverage semantic parsing methods \cite{DBLP:conf/acl/YihHM14,DBLP:conf/acl/YihRMCS16} with conventional statistical models \cite{DBLP:conf/aaai/ZelleM96,blei2003latent}. These methods heavily rely on manual annotations and predefined rules, which can hardly be transferred to other domains for further generalization. Recently, proliferated deep learning approaches \cite{DBLP:conf/aaai/ZhangDKSS18,DBLP:conf/emnlp/SunDZMSC18} enable us to train a model in an end-to-end fashion with weak supervision, and have achieved impressive performance.

\begin{figure}
	\centering  

	\includegraphics[width=0.8\linewidth]{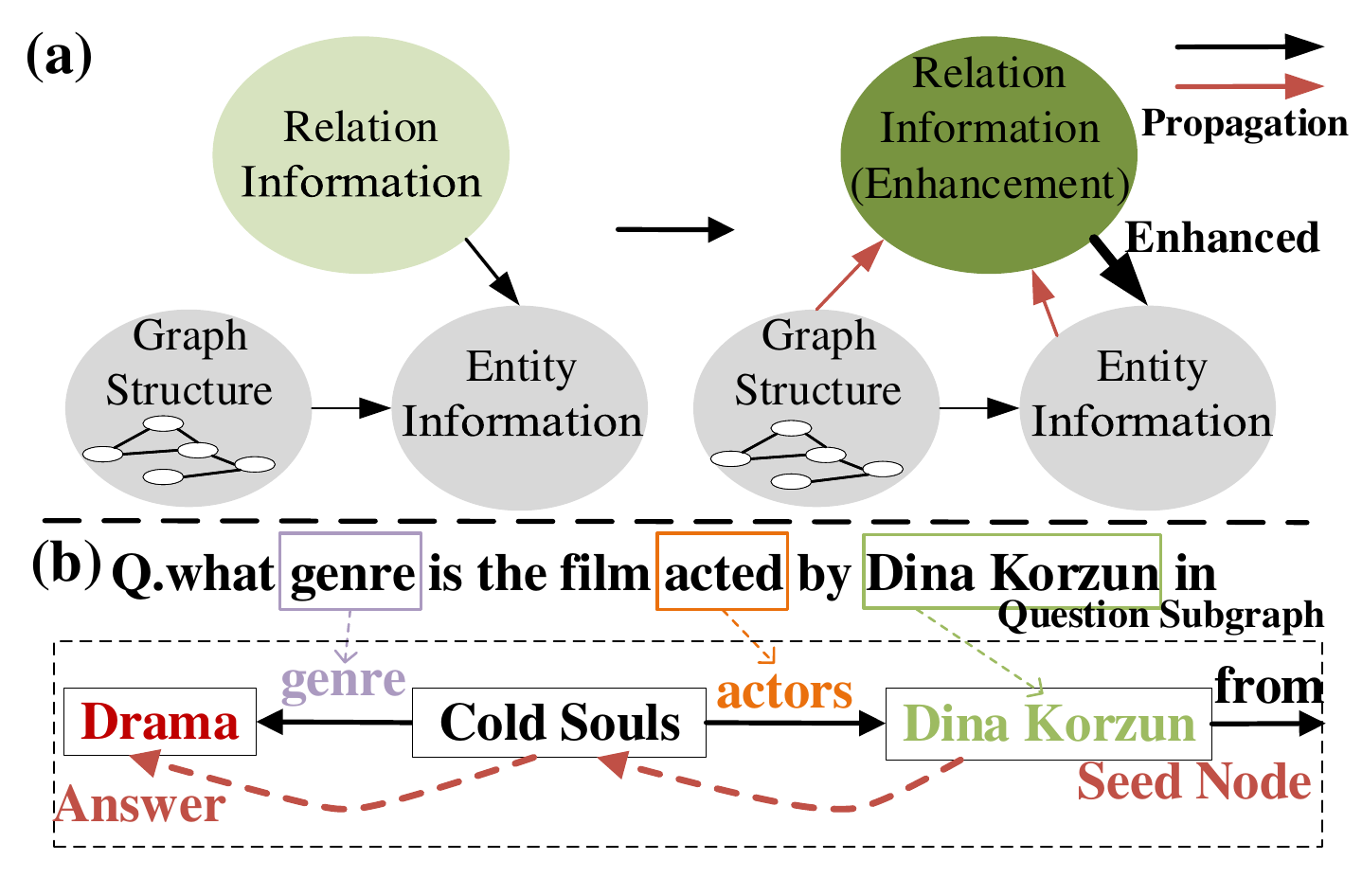}  
	\caption{\textbf{(a)}: The information flow direction of the previous approaches (left) and two new propagation flows (right) proposed are represented by the red lines.
	 \textbf{(b)}: A part of question-related subgraph used to answer the given question. Previous models make information propagate along the direction of the \textbf{black lines} to execute the reasoning process. Human behavior reasoning is in the direction of \textbf{red edges}. }
	\label{example}   
\end{figure}

A more challenging, yet practical problem for KBQA is compositional (sometimes called ``multi-hop'') reasoning, where the answers are expressed along a KB path which consists of entities and relations. Existing work \cite{DBLP:conf/www/WangJSWYCY19,DBLP:conf/acl/XiongYCGW19} learn the entity representations by extracting a question-related subgraph from a large KB with rich structured and relational information. Then, given a question, the corresponding answer can be predicted from multiple candidates, i.e., entities in the subgraph. The main problems for these models are the unlearnable relation representations and the lack of interactions between entities and relations in different subgraphs, as indicated in Figure~\ref{example}(a). Hence, the relation representations are unable to capture rich structured information of different subgraphs, reducing these models' capability to perform compositional reasoning. As shown in Section~\ref{ablation}, such interactions greatly improve the Hits@1 score \cite{DBLP:conf/emnlp/SunDZMSC18}.

In order to learn more informative entity representations, existing approaches \cite{DBLP:conf/emnlp/SunDZMSC18,DBLP:conf/sigir/LiuJHLQ16} update a node representation by the propagation along the graph and the aggregation of neighboring nodes, which can be regarded as reasoning process. However, these models simply treat the reasoning direction as arbitrary. Intuitively, inspired by human behaviors \cite{johnson2012inference}, a reasoning should be directional \cite{DBLP:conf/acl/DingZCYT19}. It starts from the seed node, \textit{i.e.,} the entity which resides in the question, and then propagates along the directional edges from the nodes closest to the seed node, onward to the more distant ones (see Figure~\ref{example}(b)). In Section~\ref{ablation}, we show that introducing such direction information can benefit the accuracy of reasoning.

In this paper, we propose Relation-updated Direction-guided Answer Selector (RDAS), a novel model that aims to tackle the aforementioned concerns for multi-hop reasoning on KBQA. To address the first challenge, inspired by Levi Graph \cite{gross2004handbook}, in each subgraph, we treat all relations as the graph nodes to facilitate interactions among entities and relations. Thus, a relation node can capture rich structured information for different subgraphs.
To address the second challenge, we first add a reverse edge between two adjacent nodes to ensure that information can be propagated bidirectionally. Then, we treat the seed node as the center, keep the edges whose directions are pointing away from the seed node (outside-directed edges), and prune the rest of edges to make the information propagate from seed nodes to external nodes.
The described two steps enable us to inject the directional information into each subgraph, improving the multi-hop reasoning. We perform detailed experiments on the open datasets MetaQA and PQL, demonstrating the superiority of the proposed model.

\section{Model} \label{1model}

\subsection{Task Definition}

\noindent Let  $\mathcal{K = (V,E,R)}$ denote a knowledge graph, where $\mathcal{V}$ is the set of entities in KB, and $\mathcal{E}$ is the set of triples (\textit{$e_{o}$, r, $e_{s}$}), where \textit{$e_{o}$}, \textit{$e_{s}$} $\in \mathcal{V}$ are entities and \textit{r} $\in$
$\mathcal{R}$ is the relation between \textit{$e_{o}$} and \textit{$e_{s}$}. Given a natural language question $\mathit{Q = (w_{1},...,w_{|Q|})}$ and its question-related subgraph $\mathcal{K_s = (V_s,E_s,R_s)} $, where $w_{i}$ denotes the $i$th word and $\mathcal{K_s} \subset \mathcal{K} $, the model needs to extract its answers from $\mathcal{V_s}$.

The rest of this section is organized as follows. Subsection~\ref{subgraph} describes how to construct subgraph-aware relations. Subsection~\ref{direction} presents the incorporating of direction information. Subsection~\ref{reasoning} introduces the multi-hop reasoning, followed by answer prediction subection~\ref{answer_prediction}.

\subsection{Subgraph-aware Relations Construction} \label{subgraph}
\begin{figure}[t]
	\centering  
	\small
	\includegraphics[width=0.85\linewidth]{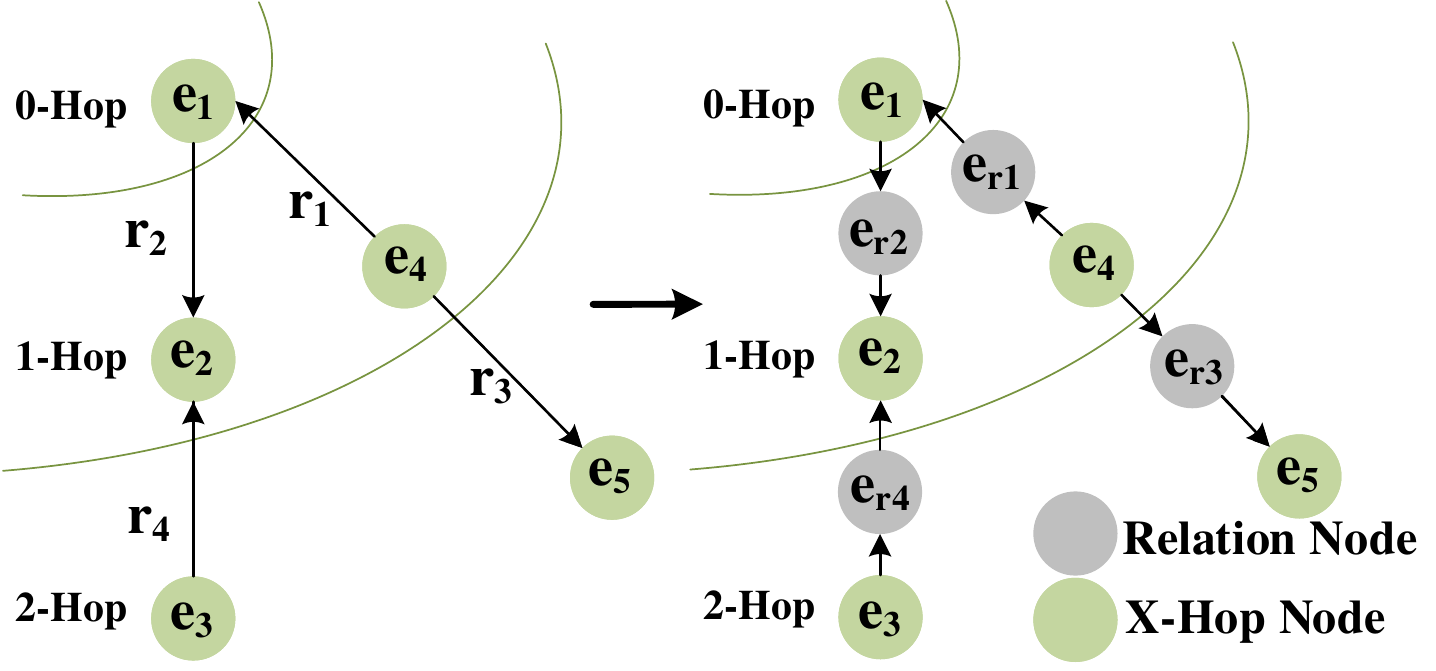}  
	\caption{Transforming relation edges into relation nodes. We regard 0-Hop nodes as seed nodes. Entity nodes are green and relation nodes are gray.}
	\label{hg_model}   
\end{figure}

To make the relation $r_{i} \in \mathcal{R_s}$ learnable and capture rich graph information, we transform the relation edges to relation nodes, defined as:
\begin{center}
	$\{e_{r_{1}},e_{r_{2}},...,e_{r_{n}}\}$,
\end{center}

\noindent where $\{r_{1}, r_{2},...,r_{n}\} = \mathcal{R_s}$. The process is shown in Figure~\ref{hg_model}. E.g., for a triple ($e_{1}, r_{2}, e_{2}$) in Figure~\ref{hg_model}, we regard $r_{2}$ as a node $e_{r_{2}}$, and add it to the middle of $e_{1}$ and $e_{2}$. It is the same as the process of Levi Graph.

From another point of view, through these steps, we construct an adjacency matrix of entities and relations. This enables relations to update their representations based on the surrounding entity information, capturing the rich structured information.

\begin{figure}[t]
	\centering  
	\small
	\includegraphics[width=0.85\linewidth]{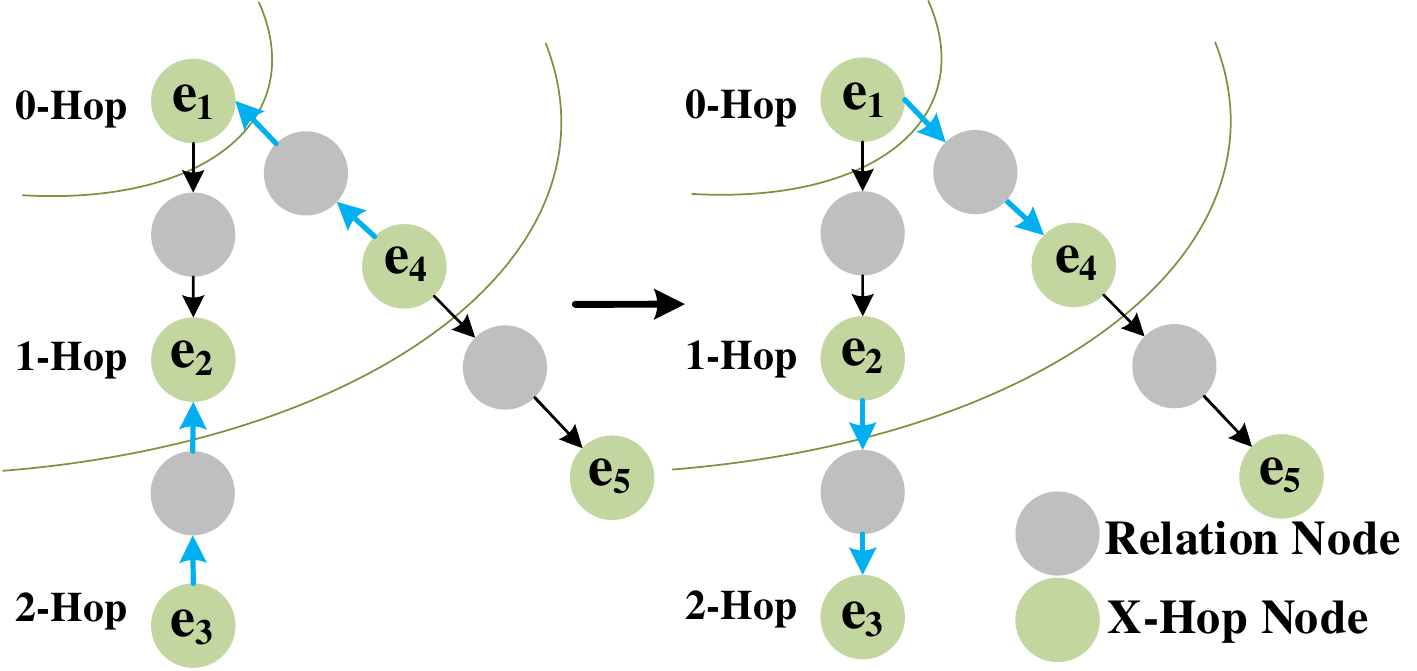}  
	\caption{The subgraph shape before and after integrating direction information. We regard the 0-Hop node as the seed node, and then we change the direction of the edges between $e_{2}$ and $e_{3}$, $e_{1}$ and $e_{4}$.}
	\label{direction_model}   
\end{figure}

\subsection{Direction Information Construction} \label{direction}

We introduce a way of integrating the direction information into the subgraph.
For each edge between two adjacent nodes, we first introduce the reverse edge to ensure that information can be propagated bidirectionally.

\noindent \textbf{Pruning}. At this step, we apply the direction information by masking some edges which is inspired by human behaviors \cite{johnson2012inference}. Similar to Figure~\ref{example}(b), we adopt a way humans search for the answers to a question. It is natural to start from the seed node (0-Hop node) ``Dina Korzun'' of the question. Secondly, the 1-Hop node ``Cold Souls" of the node ``Dina Korzun'' is picked out based on the relation ``the films acted by Dina Korzun''. Thirdly, humans find the 2-Hop node 'Drama' of the node 'Dina Korzun' based on the relation 'What genres are the films in'. Finally, the 2-Hop node ``Drama" is the answer of the question.
We adopt the way of searching for answers from the seed nodes to n-Hop nodes along the outside-directed edges.
We expect the information propagation in the model to follow such rules, so we mask the edges which are inside-directed edges. 
Figure~\ref{direction_model} shows the state of the subgraph before and after direction information construction.

\subsection{Reasoning on Graph} \label{reasoning}

After the above subsection, we get the subgraph representation $\mathcal{K}_{e}$ $= (\mathcal{V}_{e},\mathcal{E}_{e},\mathcal{R}_{e})$.

\noindent  \textbf{Node Initialization}. 
We represent all of the nodes in the graph with pre-trained word vectors, noted as $w_{v}$ $\in \mathbb{R}^{n}$ for node $e_{v}$, where $n$ is the embedding size. We also embed the distance from the node $e_{v}$ in the graph to the seed node, as $d_{v}$ $\in \mathbb{R}^{n}$. For simplicity, $d_{v}$ is represented with the embeddings of words ``0", ``1", ``2", etc. We concatenate $w_{v}$ and $d_{v}$ as the initial node representation, defined as:

\begin{equation}
n_{v}=[w_{v};d_{v}]W^{2n \times n}, n_{v}  \in \mathbb{R}^{n},
\end{equation}
where $[;]$ represents concatenation of vectors and $W^{2n \times n}$ is a learned parameter matrix.
By adding distance information, nodes can better update themselves according to the number of hops needed to infer the answers for the current question.

To represent $Q$, let $w^{Q}_{1},...,w^{Q}_{|Q|}$ be the word vectors in the question. A long short-term memory network (LSTM) \cite{DBLP:journals/neco/HochreiterS97} is used to encode the question:
\begin{equation}
q=LSTM(w^{Q}_{1},...,w^{Q}_{|Q|}),
\end{equation}
where $q \in \mathbb{R}^{m}$ is the final state from the output of the LSTM and $m$ is the hidden state size. We use $q$ to represent the question.

\noindent \textbf{Node Updates}.
To avoid question agnostic nodes, we first concatenate each node representation $n_{v}$ with the question $q$, which is defined by $h_{v}^{0}=[n_{v};q]$. 
We perform GCN on each node $e_{v}$, updated as:
\begin{equation}
u_{v}^{l+1} = \sigma \Bigg ( \sum\limits_{j\in N_{v}} \frac{1}{c_{v}} W_{1}^{l}h_{j}^{l} + W^{l}_{0}h_{v}^{l} \bigg),    
\end{equation}
where $0 \leq l<L$ and $L$ is the number of layers in the model. The $h_{v}^{l}$ donates the hidden state of node $e_{v}$ at the $l$th layer. Matrices $W_{1}^{l} \in \mathbb{R}^{d_{l+1} \times d_{l}}$ and $W_{0}^{l} \in \mathbb{R}^{d_{l+1} \times d_{l}}$ stand for learnable parameter matrices, while $d_{l}$ and $d_{l+1}$ represent hidden state dimensions of the layer $l$ and $l+1$. The $N_{v}$ represents the set of neighboring indices of the node $e_{v}$. The normalization constant $c_{v}$ can be learned or set directly, such as $c_{v}=|N_{v}|$.
The $\sigma(\cdot)$ is the sigmoid function.

A gate mechanism decides how much of the update message $u_{v}^{l+1}$ propagates to the next step.
Gate levels are computed as:
\begin{equation}
a_{v}^{l+1} = \sigma \bigg ( f_{a}\bigg ( [u_{v}^{l+1};h_{v}^{l}]\bigg) \bigg),
\end{equation}

\noindent where $f_{a}$ is a linear function. Ultimately, the next layer representation $h_{v}^{l+1}$ of the node $e_{v}$ is a gated combination of the previous representation $h_{v}^{l}$ and a non-linear transformation of the update information $u_{v}^{l+1}$:
\begin{equation}
h_{v}^{l+1} = \phi(u_{v}^{l+1}) \odot a_{v}^{l+1}+h_{v}^{l}\odot (1-a_{v}^{l+1}),
\end{equation}
where $\phi(\cdot)$ is any nonlinear function and $\odot$ stands for element-wise multiplication.

The model stacks such networks for $L$ layers. Through the convolution operation of $L$ times, the node constantly updates its own state, which simulates the reasoning process. We get the last layer representation $h_{v}^{L}$ of the node $e_{v}$ for the answer prediction.

\subsection{Answer Prediction}  \label{answer_prediction}
We transform the answer prediction problem into a binary classification problem on each node, and convert the node representation $h_{v}^{L}$ to a two-dimensional vector, defined as:
\begin{equation}
h_{v}^{p} = softmax([h_{v}^{L};q]W_{p}), h_{v}^{p}  \in \mathbb{R}^{2}, \label{eq}
\end{equation}

\noindent where $W_{p}$ is a learned parameter matrix. 

\noindent \textbf{Model Training}.
The model predicts the probability of each node (entity nodes and relation nodes) in the graph being an answer individually:

\begin{equation}
L(\theta) = -\frac{1}{m}\sum y_{v}log p_{\theta}(h_{v}^{p}), \quad v \in \mathcal{V}_{e},
\end{equation}
where $\theta$ is the model parameters, $m$ is the number of nodes in $\mathcal{V}_{e}$, and $y_{v} = [0,1]$ if the node is an answer or $y_{v} = [1,0]$ otherwise.

\section{Experiments}

\subsection{Datasets} \label{entity_linking}

\noindent  \textbf{ (1) MetaQA} \cite{DBLP:conf/aaai/ZhangDKSS18} is composed of three sets of question-answer pairs in natural language form (1-hop, 2-hop, and 3-hop) and a movie domain knowledge base.  \textbf{ (2) PQL} (PathQuestion-Large) is a multi-hop KBQA dataset \cite{DBLP:conf/coling/ZhouHZ18}. The dataset consists of 2-Hop (PQL-2H) questions and 3-Hop (PQL-3H) questions.
In order to perform entity linking, we utilize   the simple surface-level matching \cite{DBLP:conf/acl/XiongYCGW19,DBLP:conf/emnlp/SunDZMSC18} to make fair comparisons.

\subsection{Evaluation Metrics}

\noindent \textbf{Full}:
The metric is used for multi-answer prediction. For the representation $h_{v}^{p}$ of the node $e_{v}$, as shown in Equation~\ref{eq}, if the second dimension value is higher than the first dimension value, the node is regarded as an answer. For a question, if the predicted answer set is the same as the gold answer set, we set Full=1, else Full = 0. We average Full on the test set to get the final score.

\noindent \textbf{Hits@1}:
The model selects the node with the largest second dimension value as the final answer. If this answer is in the gold answer list, we regrad the prediction as correct. We average the correct on the test set to get the final score.

\begin{table}[t]
    \centering
    \setlength{\tabcolsep}{0.4mm} {
			\begin{tabular}{lcc|cc|cc}
				\toprule
				
				\multirow{3}*{Model} 
				&\multicolumn{2}{c|}{MetaQA 1-Hop} &\multicolumn{2}{c}{MetaQA 2-Hop} &\multicolumn{2}{|c}{MetaQA 3-Hop}\\
				\cline{2-7}

				&Hits@1 & Full &Hits@1 & Full &Hits@1 & Full \\
				\hline  
				KVMem & 0.958 & 0.890 &  0.760 & 0.643 & 0.489 & 0.173 \\ 
				VRN & 0.978 & 0.895 &  0.898 & 0.720 &  0.630 & 0.250  \\ 
				SGReader & 0.967 & 0.903 & 0.807 & 0.719 &  0.610 & 0.272  \\ 
				GraftNet & 0.974 & 0.918 &0.950 & 0.681 &  0.778 & 0.226 \\ 
				\bf RDAS & \textbf{0.991} & \textbf{0.976} &  \textbf{0.970} & \textbf{0.802} &  \textbf{0.856} & \textbf{0.275}  \\ 
				\hline
				\hline
				VRN*  & 0.975 & -  & 0.899 & -  & 0.625 & -   \\
				GraftNet*  & 0.970 & -   & 0.948 & -  & 0.777 & -  \\
			
				\bottomrule
			\end{tabular}
		}
    \caption{\label{table_metaQA}Experimental results on the MetaQA datasets. The symbol * indicates the numbers are from the original papers.}
    \label{MetaQA}
\end{table}

\begin{table}[t]
    \centering
    \setlength{\tabcolsep}{2mm}{
		
			\begin{tabular}{lcc|cc}
				\toprule
				
				\multirow{2}*{Model} 
				&\multicolumn{2}{c|}{PQL-2H} &\multicolumn{2}{c}{PQL-3H}\\
				\cline{2-5}
				&Hits@1 & Full &Hits@1 & Full \\
				\hline   
				KVMem & 0.622 & 0.450 & 0.674 & 0.689 \\ 
				IRN & 0.725 & 0.590 & 0.710 & 0.801 \\  
				SGReader & 0.719 & 0.626 & 0.893 & 0.825  \\  
				GraftNet & 0.706 & 0.269 & \textbf{0.913} & 0.408  \\  
				\hline   
				\bf RDAS & \textbf{0.736} & \textbf{0.691} & 0.910 & \textbf{0.861}  \\  
				\bottomrule  
			\end{tabular}
	}
    \caption{\label{table_PQL}Experimental results on the PQL datasets.}
    \label{PQL}
\end{table}

\subsection{Baselines}
\noindent We use five baseline methods for comparison, including KVMem \cite{DBLP:conf/emnlp/MillerFDKBW16}, IRN \cite{DBLP:conf/coling/ZhouHZ18}, 
VRN \cite{DBLP:conf/aaai/ZhangDKSS18}, GraftNet \cite{DBLP:conf/emnlp/SunDZMSC18} and SGReader \cite{DBLP:conf/acl/XiongYCGW19}.

\subsection{Main Results and Discussion}

\noindent  The experimental results on the MetaQA are shown in Table~\ref{table_metaQA}. From the results, we can observe that:

(1) For the Hits@1 metric , our model achieves best or competitive results. For the 2-Hop and the 3-Hop subsets, our model performance increases by 2.0\% and 7.8\%, respectively. SGReader and GraftNet, which use the GCN to perform reasoning,  ignore the encoding of subgraph-aware relation information and have no direction information to guide themselves. This might be the reason for our model's better performance.

(2) Compared with the single answer prediction setting (Hits@1), mutli-answer prediction is more challenging. Still, according to the Full metric, our model yields significant improvements. This suggests that the integration of graph-aware relation representations and direction information improves the performance in predicting multi-answer.

We also compare our model with baselines on the PQL dataset. Although it is a single-answer dataset, we still use it to showcase our model's ability to predict answers on each node. The experimental results on PQL are shown in Table~\ref{table_PQL}. 
From the results, we observe that our model achieves competitive results for the Full metric compared with the Hits@1, which proves the validity of our model in the node prediction.

\begin{table}
	\begin{center}
		\setlength{\tabcolsep}{1.5mm}{
			
			\begin{tabular}{lcccc}
				\toprule
				
				\multirow{2}*{Model} 
				&\multicolumn{2}{c}{MetaQA 2-Hop} &\multicolumn{2}{c}{PQL-3H}\\
				\cline{2-5}
				&Hits@1&Full&Hits@1&Full\\
				\hline
				\bf RDAS&\textbf{0.970}&\textbf{0.802}&\textbf{0.910}&\textbf{0.861}\\
				\hline
				No RN&0.937&0.766&0.887&0.790\\
				No Direction&0.942&0.740&0.875&0.788\\
				No DE&0.966&0.784&0.857&0.808\\
				
				\bottomrule
			\end{tabular}
		}
	\end{center}
	 \vspace{-0.5cm}
	\caption{\label{Ablation Experiment}Ablation experiments of RDAS.}
\end{table}

\subsection{Ablation Experiment of RDAS} \label{ablation}

\noindent To study the contributions of the main components in the RDAS, we conduct ablation experiments. The results are shown in Table~\ref{Ablation Experiment}, where (No RN) means removing relations nodes, (No Direction) expresses removing direction information, and (No DE) means removing distance embedding, respectively.  Overall, we observe that removing each component from RDAS will lead to a significantly Hits@1 drop and Full drop on both datasets, which indicates the effectiveness of each component.

\section{Conclusion}

\noindent  In this paper, we devote ourselves to solving single-answer and multi-answer prediction tasks, and propose a novel model, named Relation-updated Direction-guided Answer Selector (RDAS). 
The proposed model utilizes subgraph-aware relation representations for capturing rich structured information and introduces the direction information into the graph to enhance the reasoning ability. Experiments based on two open datasets demonstrate our model's ability.

\section*{Acknowledgements}
This work is supported by Beijing Nova Program of Science and Technology (Grant No. Z191100001119031), National Key Research and Development Program of China (Grant No. 2018YFB1003804), and The Open Program of Zhejiang Lab (Grant No. 2019KE0AB03). Xu Wang is supported by  BUPT Excellent Ph.D. Students Foundation under grant CX2019137. Shuai Zhao is the corresponding author.

\bibliographystyle{IEEEbib}
\bibliography{strings,refs}

\end{document}